\documentclass[final,5p,times,twocolumn,authoryear]{elsarticle}

\usepackage{amssymb}
\usepackage{amsmath}

\usepackage{soul}
\usepackage{color}
\usepackage{hyperref}
\usepackage{comment}
\usepackage{subcaption}
\usepackage{multirow}
\usepackage{makecell}
\usepackage{tikz}
\usetikzlibrary{arrows.meta,positioning,calc,fit,shapes.misc,decorations.pathmorphing}

\journal{TBD}

\begin{document}
	
	\begin{frontmatter}
		
		
		
		\title{SurgPhase: Time efficient pituitary tumor surgery phase recognition via an interactive web platform} 
			\author[1]{Yan Meng\corref{cor1}} \cortext[cor1]{Corresponding author} \fnref{label1} \ead{ ymeng@childrensnational.org}
			\author[2]{Jack Cook\fnref{label1} } 
			\author[3]{X.Y. Han}
			\author[2]{Kaan Duman}
			\author[2]{Shauna Otto }
			\author[10]{Dhiraj Pangal}
			\author[4,5]{Jonathan Chainey}
			\author[6,7]{Ruth Lau}
			\author[2]{Margaux Masson-Forsythe}
			\author[1,2]{Daniel A. Donoho}
			\author[8]{Danielle Levy}
			\author[8]{Gabriel Zada}
			\author[9]{Sébastien Froelich}
			\author[10]{Juan Fernandez-Miranda}
			\author[10]{Mike Chang}
			
			\fntext[label2]{Yan Meng and Jack Cook contributed equally to this work.}
			\affiliation[1]{organization={Children's National Hospital},
				addressline={111 Michigan Ave NW}, 
				city={Washington},
				postcode={20010}, 
				state={DC},
				country={USA}}
			\affiliation[2]{organization={Surgical Data Science Collective},
				addressline={2000 Pennsylvania Ave. NW, Suite 7000}, 
				city={Washington},
				postcode={20006}, 
				state={DC},
				country={USA}}
			\affiliation[3]{organization={Booth School of Business, University of Chicago},
				addressline={5807 S Woodlawn Ave}, 
				city={Chicago},
				postcode={60637}, 
				state={IL},
				country={USA}}
			\affiliation[4]{organization={Division of Neurosurgery, Department of Surgery, University of Montreal},
				addressline={2900 Edouard Montpetit Blvd}, 
				city={Montreal},
				postcode={H3T 1J4}, 
				state={IL},
				country={Canada}} 
			\affiliation[5]{organization={Wilson Centre for Research in Education, University Health Network},
				addressline={190 Elizabeth St}, 
				city={Toronto},
				postcode={M5G 2C4}, 
				state={ON},
				country={Canada}}
			\affiliation[6]{organization={Department of Neurosurgery, Hospital Universitari Joan XXIII},
				addressline={Carrer Dr. Mallafre Guasch}, 
				city={Tarragona},
				postcode={43005}, 
				country={Spain}}
			\affiliation[7]{organization={Universitat Rovira i Virgili},
				addressline={Carrer de l'Escorxador}, 
				city={Tarragona},
				postcode={43003}, 
				country={Spain}}
			\affiliation[8]{organization={University of Southern California},
				addressline={ 3501 S Figueroa St}, 
				city={Los Angeles},
				postcode={43003}, 
				state={CA},
				country={USA}}     
			\affiliation[9]{organization={Hôpital Lariboisière},
				addressline={ 2 Rue Ambroise Paré}, 
				city={Paris},
				postcode={75010 }, 
				country={France}} 
			\affiliation[10]{organization={ Stanford University School of Medicine},
				addressline={300 Pasteur Drive}, 
				city={Stanford},
				postcode={94305}, 
				state={CA},
				country={USA}}                	
			
			\begin{abstract}
				Accurate surgical phase recognition is essential for analyzing procedural workflows, supporting intraoperative decision-making, and enabling data-driven improvements in surgical education and performance evaluation. In this work, we present a comprehensive framework for phase recognition in pituitary tumor surgery (PTS) videos, combining self-supervised representation learning, robust temporal modeling, and scalable data annotation strategies. Our method achieves 90\% accuracy on a held-out test set, outperforming current state-of-the-art approaches and demonstrating strong generalization across variable surgical cases.
				
				A central contribution of this work is the integration of a collaborative online platform designed for surgeons to upload surgical videos, receive automated phase analysis, and contribute to a growing dataset. This platform not only facilitates large-scale data collection but also fosters knowledge sharing and continuous model improvement. To address the challenge of limited labeled data, we pretrain a ResNet-50 model using the self-supervised framework on 251 unlabeled PTS videos, enabling the extraction of high-quality feature representations. Fine-tuning is performed on a labeled dataset of 81 procedures using a modified training regime that incorporates focal loss, gradual layer unfreezing, and dynamic sampling to address class imbalance and procedural variability.
				
				To generate training labels with minimal manual effort, we introduce a novel pipeline that automatically extracts coarse phase annotations from surgeon-provided procedure notes. Per-frame embeddings are extracted and processed using a modified Multi-Stage Temporal Convolutional Network (MS-TCN++) with enhanced feature dimensionality for temporal segmentation. Overall, this work demonstrates a practical and scalable solution for surgical phase recognition and establishes a foundation for collaborative, AI-driven surgical intelligence in neurosurgery.

			\end{abstract}

			
		\end{frontmatter}
		
			

			\section{Introduction}
			\label{sec:intro}
			Automated recognition of surgical phases from intraoperative video is a cornerstone capability for next-generation surgical intelligence systems, enabling intraoperative decision support, streamlined documentation, and enhanced surgical training and quality assurance. In endoscopic pituitary tumor surgery (PTS), robust phase recognition is particularly critical due to the confined operative corridor, complex anatomy, and proximity to vital neurovascular structures, where even minor errors can have severe consequences.
			
\begin{figure}[!t]
				\centering
				\includegraphics[width=\columnwidth]{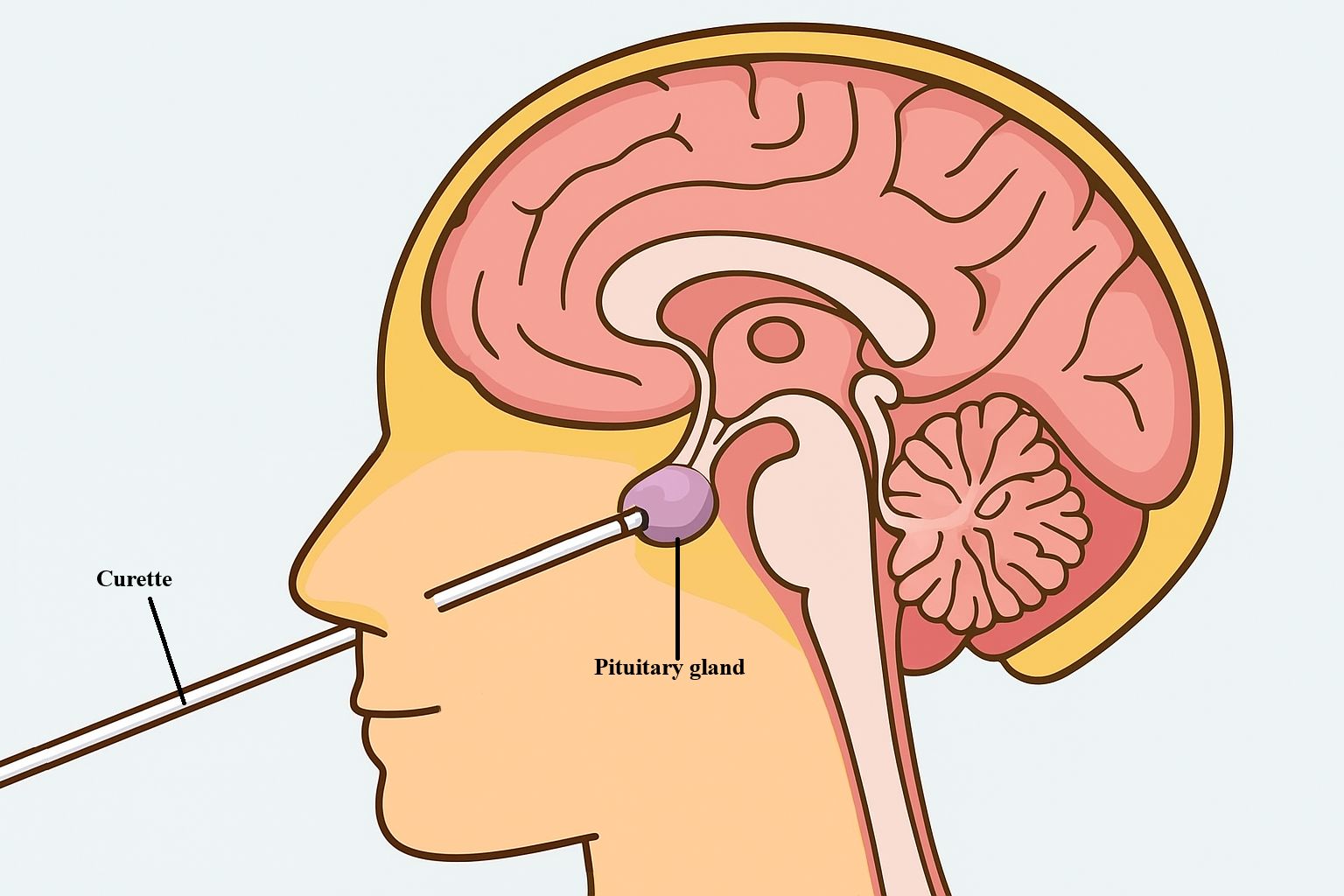}	
				\caption{Schematic overview of endoscopic transsphenoidal pituitary tumor surgery.}
				\label{fig:pts_illustration}
			\end{figure}

			Nowadays, endoscopic transsphenoidal approach (eTSA) is the minimally invasive standard for accessing and resecting pituitary tumors. The surgical pathway proceeds sequentially through four anatomical stages: nasal, sphenoidal, sellar and closure. each marked by key procedural steps and anatomical landmarks  Beginning in the nasal cavity, decongestion and flap creation optimize access. A posterior septectomy is performed to establish a binarial corridor, facilitating bimanual instrumentation and enhanced visualization via high-definition endoscopy and neuronavigation assistance. Once the sphenoid ostium is identified and opened, the anterior wall of the sphenoid sinus is removed to expose the sella turcica. Careful drilling of sphenoid septations is required to avoid vascular injury, particularly those attached to the internal carotid arteries.  After exposing the sellar floor, the dura is incised under doppler guidance to avoid harming adjacent neurovascular structures \citep{fagan2014open}. Tumor resection is typically performed in piecemeal fashion: initial debulking is followed by systematic removal of basal, lateral, and superior portions, often utilizing angled endoscopes, curettes, and suction instruments. Preservation of the normal pituitary gland and diaphragma sellae is paramount. Techniques may include bimanual dissection or double-suction methods to gradually collapse and remove suprasellar tumor, with precise oriented resection under angled optics. At the conclusion, angled endoscopy is used to inspect any remnant, with maneuvers like Valsalva tests to assess cerebrospinal fluid (CSF) integrity \citep{sharma2016endoscopic}. Reconstruction of the skull base may involve multilayered repairs employing fat grafts, fascia, glues, and nasoseptal flaps, depending on whether intraoperative CSF leak occurred. A schematic overview of the operation and anatomical landmarks is presented in Figure \ref{fig:pts_illustration}, illustrating the narrow transnasal access corridor, sphenoid sinus exposure, and the sellar region where tumor resection is performed.
			
			This technique provides direct access to the pituitary gland via a natural corridor, avoiding craniotomy, external scars, and brain retraction. Moreover, it enables high-resolution visualization in a compact field, while leveraging binocular neuronavigation and optical Doppler for safety. Multidisciplinary teams collaborate throughout the planning and execution phases to optimize outcomes , typically involving neurosurgeons, otorhinolaryngologists, endocrinologists, and radiologists. The procedure is characterized by extended duration, high inter-surgeon variability, diverse instrumentation, and fluctuating video quality. Manual annotation of such surgical phases is both labor-intensive and inconsistent across cases, limiting its scalability for research and clinical integration. Accurate temporal segmentation of PTS videos would allow for objective performance metrics, context-aware intraoperative alerts, automated operative note generation, and targeted educational review, which has the potential to materially improve patient safety and training quality. 
			
			Beyond annotation efficiency, robust phase recognition models and their derivatives have gradually improved across varied surgical domains by refining predictions across stages while preserving long-range temporal context \citep{czempiel2020tecno,yang2024surgformer}. Self-supervised learning (SSL), especially contrastive frameworks like SimCLR \citep{chen2020simple}, has further proven instrumental in learning high-quality visual embeddings from unlabeled surgical videos, reducing reliance on hand-annotated data and improving downstream tasks like phase recognition. Recent studies instructively show that SSL yields 7-14\% gains over conventional training regimes, particularly in domains where annotation is costly and limited \citep{ramesh2023dissecting}.
			
			Despite the clear clinical and societal need accompany the development of intelligent surgical workflow recognition technologies, reliable automated analysis for pituitary procedures faces two interrelated barriers. First, pituitary surgery videos pose unique visual challenges: frequent camera occlusions, small fields of view with black-bordered endoscopic frames, rapid instrument switching, and variable illumination and lens artifacts. These factors increase intra- and inter-case variability and complicate feature extraction. Second, there are very few publicly available, well-annotated corpora for endoscopic pituitary workflows; recent initiatives PitVis-2023 Challenge \citep{das2025pitvis} provide promising but still limited data resource footprints compared with larger datasets available for other procedures such as cholecystectomy. The scarcity of comprehensive annotated data restricts supervised learning approaches and slows progress on clinically relevant models.
			
			From a global perspective, the broader need for such systems is underscored by the global surgical workforce crisis beyond high-resource centers. An estimated five billion people lack access to safe, timely, and affordable surgical care, with the shortage of neurosurgical expertise being especially severe in low- and middle-income countries (LMICs) \citep{meara2015global,alkire2015global}. A cloud-based automated analysis system has the potential to bridge this gap by supporting remote mentorship, enabling asynchronous case review, and delivering AI-assisted surgical education at scale. In high-resource settings, they can also enhance operative efficiency by streamlining postoperative video analysis, a task that can otherwise consume substantial clinical time. For medical education, automated phase analysis can support case-based learning, facilitate self-assessment for trainees, and allow senior surgeons to rapidly review key operative segments \citep{hashimoto2018artificial}.
			
			To address these challenges, we present an integrated framework that leverages large-scale in-domain self-supervised pre-training, imbalance-aware fine-tuning, and high-capacity temporal modeling, all deployed within a clinician-facing platform that supports weak annotation extraction and continual data accrual. The interactive web Surgical Video Platform (SVP), developed at Surgical Data Science Collective (SDSC), connects surgeons, scientists, and engineers to unlock the insights hidden within surgical data. From a clinical and societal perspective, embedding these technical innovations within an online collaborative platform promises substantial downstream impact. Such a platform allows surgeons to upload and share videos, receive automated phase segmentation, and contribute granular notes, which facilitate continual learning, democratize access to analytics, and supporting scalable improvements in training, quality assessment, and operational efficiency. Early evidence suggests that even simple computer vision based tools can reduce the time required for surgical video review by over 70\%, unlocking time savings and enabling teamwide situational awareness during operations \cite{wong2024collaborative}. In pituitary surgery, where anatomy is delicate and errors can have severe consequences, such systems stand to directly improve patient safety, enhance educational value, and inform procedural standardization through shared analyses.
			
			Building on these foundational insights, this work introduces a holistic framework for surgical phase recognition in endoscopic pituitary surgery centered around four key contributions:
			\begin{itemize}
				\item  SSL-based pre-training leveraging large unlabeled surgical video corpora to learn robust, transferrable embeddings with limited labeled data.
				\item  Imbalance-aware fine-tuning, deploying focal loss, dynamic sampling, and gradual unfreezing to improve recognition of rare or brief surgical phases.
				\item Temporally enhanced segmentation modeling, adapting MS-TCN++ architectures to accept high-dimensional embeddings for precise phase delineation.
				\item Clinician-oriented platform deployment, enabling weak annotation via operative notes, continuous model refinement, and accessible feedback mechanisms for stakeholders.
			\end{itemize}
			
			Through these contributions, we aim to bridge technical advances with clinical utility, providing a scalable and socially impactful AI tool that supports surgical practice, ultimately fostering improved outcomes, distributed knowledge, more effective training, and equitable access to surgical expertise worldwide.
			
			\section{Related Work}
			\label{sec:related}
			
			\subsection{Surgical Phase Recognition } 
			\label{sec:surgicalphaserec}
			Early deep approaches treated phase recognition as a sequence labeling task using frame-level convolutional neural network (CNN) features followed by temporal models such as Hidden Markov model (HMM), long-short term memory (LSTM) or temporal conditional random field (CRF) \citep{twinanda2016single}. These single-stage pipelines established baseline viability and demonstrated that incorporating temporal context materially improves frame-level accuracy compared with spatial models alone. Multi-task variants that jointly predict tools and phases were also proposed to exploit correlated signals such as tool presence and anatomy for better temporal consistency \citep{ramesh2021multi}.
			
			Work from the general video community introduced end-to-end spatio-temporal feature extractors, such as I3D and R(2+1)D \citep{carreira2017quo,tran2018closer}, and they were later adapted to surgical video tasks; they are particularly useful when motion cues and short-term temporal dynamics matter. Two-stream and multi-rate designs \citep{feichtenhofer2019slowfast} help capture both slow semantic context and fast motion, which is a desirable property for videos with rapid instrument manipulations. These backbones are frequently used as the spatial front end in two-stage surgical pipelines.
			
			Multi-stage temporal convolutional networks such as MS-TCN and its derivatives \citep{lea2017temporal,farha2019ms,li2020ms,czempiel2020tecno} became a dominant family for long-range action segmentation because they balance receptive field size and computational efficiency while reducing over-segmentation via iterative refinement. These architectures are widely used in surgical phase segmentation benchmarks and form the backbone of many strong baselines \citep{park2022multi,fang2022global}. Variants incorporate moment losses, global-local aggregation, or multi-stream inputs including instrument and anatomy, to handle intraphase heterogeneity.
		
			Following the success of transformer architectures and attention mechanisms \citep{vaswani2017attention} for natural language processing in 2020-2022, recent works --- such as ASFormer \citep{yi2021asformer}, OperA \citep{czempiel2021opera}, Trans-SVNet \citep{gao2021trans}, LoViT \citep{liu2025lovit}, MuST \citep{perez2024must}, and SurgFormer \citep{yang2024surgformer,meng2025ai} --- built transformer-based pipeline for action segmentation and surgical phase recognition. They demonstrate that innovations such as multi-scale temporal self-attention \citep{perez2024must}, hierarchical attention \citep{yang2024surgformer}, and hybrid spatial-temporal Transformer \citep{gao2021trans} can achieve or exceed prior state-of-the-art on standard benchmarks. However, the quadratic computational and memory complexity of standard self-attention induces a major practical challenge in their implementation. To address this, the above architectures leverage frame downsampling and/or local-truncation (such as clipping or sliding window attention) techniques to conform to time, cost, and hardware constraints. However, in our online surgical phase recognition platform --- where surgeons upload videos and expect prompt analytic feedback --- we found that transformer-based models (even with the downsampling and local-trunction) induce inference latency and large memory footprints that are too high and impractical. Instead, we found that our MS-TCN++ pipeline induces much shorter latency and allows us to use much less aggressive frame downsampling schemes than transformer-based solutions: The latter benefit also helps preserve more fine-grained temporal information.

			\subsection{Self-supervised Pre-training}
			The scarcity of labeled surgical video has motivated extensive use of pre-training and self-supervised learning. Generic contrastive frameworks \citep{chen2020simple} and domain-specific pre-training efforts \citep{batic2024endovit} have produced representations that improve downstream phase and scene tasks when labeled data are limited. Domain-specific pretraining often outperforms out-of-domain ImageNet initialization for endoscopy tasks. These findings motivate us using large corpora of unlabeled pituitary videos for SSL pre-training to improve phase recognition. 
			
			\subsection{Surgical Benchmarks and Datasets}
			Progress in surgical phase recognition has been driven by curated public benchmarks including Cholec80 \citep{endonet}, M2CAI challenge \citep{twinanda2016single}, EndoVis \citep{nwoye2023cholectriplet2021} and CATARACTS \citep{al2019cataracts}, which provide standardized tasks for phase, tool and scene understanding and have enabled rapid method comparison. These datasets remain concentrated in abdominal and ophthalmic procedures, and are substantially larger and more systematically annotated than most skull-base pituitary collections, a disparity that complicates direct transfer of models to endoscopic pituitary workflows. 
			
			\subsection{Annotation Strategies}
			Manual frame-level annotation is the critical bottleneck for long-form surgical video. A growing literature explores weak supervision, such as with video-level tags, tool presence, operation notes, and automated captions; semi-supervised learning, and automated extraction of coarse labels from operative notes or speech transcripts are also used to reduce annotation cost \citep{nyangoh2023systematic}. Several pituitary-focused works have already shown that workflow recognition can support automatic operation-note generation and that weak/auto annotations materially accelerate dataset curation. Systematic reviews highlight annotation heterogeneity and propose standardized semantics for surgical process modeling, which reinforces the value of pipelines that can ingest clinician notes to generate usable weak labels \citep{das2023automatic}.
			
			Taken together, the literature shows promising development in video temporal modeling and their application in medical field. However, pituitary and general endoscopic skull-base surgery remains relatively under-represented in public datasets and published benchmarks; the domain’s small field of view, frequent occlusions and variable recording conventions demand method adaptations. These domain specific needs motivate us to close the translational gap for pituitary surgical phase recognition.

			\section{Methods}
			\label{sec:method}
			This section describes our proposed pipeline for surgical phase recognition in pituitary tumor surgery (PTS) videos. The system comprises three main stages: (1) self-supervised pre-training of a visual encoder, (2) supervised fine-tuning with temporal modeling, and (3) collaborative data collection and automated annotation via an online platform.
			
			\subsection{Self-Supervised Pre-training}
			\label{sec:ssp}

			To exploit the abundance of unlabeled surgical video data, we adopt SimCLR \citep{chen2020simple}, a contrastive self-supervised learning (SSL) framework, to pretrain a ResNet-50 encoder $f_{\theta}$ on 251 unlabeled PTS procedures uploaded to the platform by surgeons from multiple institutions. SimCLR encourages the encoder to produce similar embeddings for different augmentations of the same frame while contrasting them against embeddings from different frames. For two augmented views $x_{i}$ and $x_j$ of the same input $x$, the contrastive loss is defined as Equation \ref{eq:contrast_loss}.
			
			\begin{equation}
				\mathcal{L}_{i,j} = -\log \frac{\exp(\text{sim}(z_i, z_j)/\tau)}{\sum_{k=1}^{2N} \mathbf{1}_{[k \ne i]} \exp(\text{sim}(z_i, z_k)/\tau)}
				\label{eq:contrast_loss}
			\end{equation}
			
			\noindent where $z_i = g(f_\theta(x_i))$ is the projection of the encoder output, $\text{sim}(a,b) = a^\top b / \|a\| \|b\| $ denotes cosine similarity, $\tau$ is a temperature parameter, $N$ is the batch size, and $\mathbf{1}_{[k \ne i]} $ is a sign function that is true iff  $k \ne i$.
			
			\subsection{Automated Annotation Pipeline}
			Given the variability in start and end points across procedures, manual annotation is time-consuming and error-prone. To address this, we implemented a NLP-based strategy that extracts coarse phase annotations from surgeon-provided operative notes. On our SVP user interface, surgeons are able to leave HH:MM:SS timestamps with notes including phase boundaries. We then convert that to seconds and using the original and converted frame rate, we can find the frames at which the phase transitions occur. These annotations serve as weak supervision to accelerate training data preparation and reduce labeling latency.
			
			\subsection{Supervised Fine-Tuning with Balanced Learning}
			\label{sec:sft}
			
			After the self-supervised pre-training stage, the ResNet-50 encoder is fine-tuned on a carefully curated and manually labeled surgical procedures \citep{he2016deep}. A key challenge during supervised training is the significant class imbalance across different surgical phases. Some phases, such as tumor resection in the sellar, occupy a much larger proportion of the video timeline compared to brief transitional stages like nasal access or closure. To mitigate this imbalance, we replace the standard cross-entropy loss with focal loss \citep{lin2017focal}, as defined in Equation~\ref{eq:focal}. Focal loss dynamically scales the loss contribution of each sample based on its predicted confidence, thereby reducing the relative impact of well-classified (majority class) examples and focusing the model's learning on harder, underrepresented class samples. This results in improved sensitivity to short or infrequent surgical phases, which are critical for accurate procedural segmentation.
			
			\begin{equation}
				\mathcal{L}_{\text{focal}}(p_t) = -\alpha_t (1 - p_t)^\gamma \log(p_t)
				\label{eq:focal}
			\end{equation}
			
			\noindent where $p_t$ is the model's predicted probability for the true class, $\alpha_t \in [0,1]$ is a weighting factor that balances the importance of different classes, and $\gamma$ is the focusing parameter adjusting the rate at which easy examples are down-weighted, it is set empirically. 
			
			Furthermore, to preserve the pretrained features and stabilize convergence during transfer learning, we employ a gradual layer unfreezing strategy, where layers of the ResNet-50 backbone are incrementally unfrozen across pre-determined training epochs.
			
			\subsection{Temporal Modeling with MS-TCN++}
			
			To model the temporal structure of surgical workflows, we adopt MS-TCN++ \citep{farha2019ms}, a multi-stage temporal convolutional network architecture designed for action segmentation. Unlike recurrent architectures, MS-TCN++ avoids sequential dependencies in computation, enabling efficient parallelization. Moreover, the multi-stage refinement progressively corrects misclassifications and smooth phase boundaries, which is critical in pituitary tumor surgery where subtle anatomical changes signal phase transitions. While transformer-based models \citep{gao2021trans} can capture complex temporal relationships, their quadratic complexity in sequence length and higher inference latency make them less suitable for our online deployment requirements (see discussion in Section \ref{sec:surgicalphaserec}).
			
			We remove the ResNet-50 classification head and extract per-frame embeddings for each time step $t$. These embeddings are stacked and passed to the MS-TCN++ model. Let $ \mathbf{X} = \{\mathbf{x_t} \in \mathbb{R}^d | t=1,...,T\}$ denote the sequence of $d$-dimensional feature vectors for $T$ video frames. These features are passed to the first stage of the multi-stage temporal convolutional network, which employs multi-dilated temporal convolutions to capture dependencies across varying temporal ranges \citep{khalfaoui2021dilated}. The standard MS-TCN++ input channels are increased from 64 to 256 in our architecture to accommodate richer embedding representations. The model consists of multiple temporal convolutional blocks with dilation to capture long-range dependencies across surgical phases. Each stage $s \in \{1,...,S\}$ refines the phase predictions from the previous stage. For the first stage, the hidden activations are computed as in Equation \ref{eq:h}.
			
			\begin{equation}
				\mathbf{H}^{(1)} = f_{\text{TCN}_1}(\mathbf{X})
				\label{eq:h}
			\end{equation}

			\noindent where $f_{\text{TCN}_1}$ denotes the temporal convolutional blocks with dual dilation kernels of low and hight dilation factors respectively. The general operations in the dual dilated layer are described in Equation \ref{eq:conv}, where $\mathbf{W}_{d_{1}}$,  $\mathbf{W}_{d_{2}}$ are the wights of dilated convolutions, and $\mathbf{b}_{d_{1}}$, $\mathbf{b}_{d_{2}}$ are the bias vectors.
			
			\begin{equation}
				f_{\text{TCN}_{s,l}}=\mathbf{H}^{s}_{l-1}+\mathbf{W} * ReLU(\mathbf{W}_{d_{1}}*\mathbf{H}^{s}_{l-1}+\mathbf{b}_{d_{1}} + \mathbf{W}_{d_{2}}*\mathbf{H}^{s}_{l-1}+\mathbf{b}_{d_{2}}) + \mathbf{b}
				\label{eq:conv}
			\end{equation}
			
			A $1 \times 1$ convolution layer is applied after the last dilated convolution layer, and the corresponding phase probabilities $\mathbf{P} $ are obtained via a softmax layer in Equation \ref{eq:single_pred}. 
			
			\begin{equation}
				\mathbf{P}^{(1)} = \mathrm{softmax}(\mathbf{W}^{(1)} \mathbf{H}^{(1)} + \mathbf{b}^{(1)})
				\label{eq:single_pred}
			\end{equation}
			
			
			For subsequent stages $s>1$, the input is the probabilities from the previous stage, enabling iterative refinement.
			
			\begin{equation}
				\mathbf{H}^{(s)} = f_{\text{TCN}_s}(\mathbf{P}^{(s-1)}), \quad
				\mathbf{P}^{(s)} = \mathrm{softmax}(\mathbf{W}^{(s)} \mathbf{H}^{(s)} + \mathbf{b}^{(s)})
			\end{equation}
			
\begin{figure*}[!t]
				\centering
				\includegraphics[width=0.7\columnwidth]{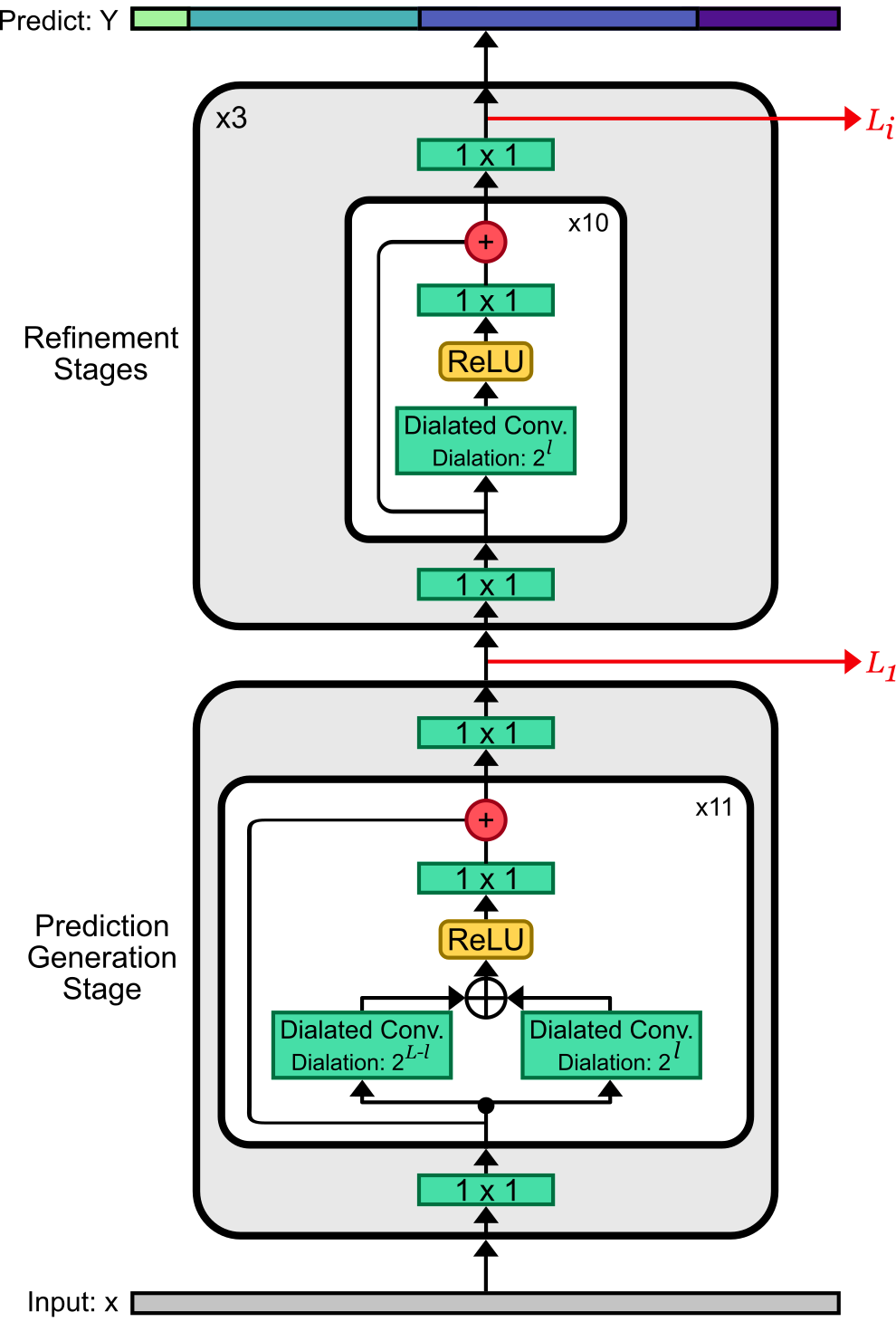}	
				\caption{Improved MS-TCN++ architecture.}
				\label{fig:mstcn}
			\end{figure*}

			Instead of the conventional cross-entropy, focal loss in Equation \ref{eq:focal} is again applied during temporal model training to address phase imbalance in surgical datasets. We further regularize the loss function with a temporal smoothing loss to penalize abrupt phase transitions as shown in Equation \ref{eq:smooth_loss}, hence the overall training objective across all $S$ stages is in Equation \ref{eq:total_loss}. The improved MS-TCN architecture is shown in Figure \ref{fig:mstcn}.
			
			\begin{equation}
				\mathcal{L}_{\text{smooth}} = \frac{1}{T \cdot C} \sum_{t=2}^T \sum_{c=1}^C \left| \log p_{t,c} - \log p_{t-1,c} \right|
				\label{eq:smooth_loss}
			\end{equation}
			
			\begin{equation}
				\mathcal{L} = \sum_{s=1}^S \left( \mathcal{L}_{\text{focal}}^{(s)} + \lambda \mathcal{L}_{\text{smooth}}^{(s)} \right)
				\label{eq:total_loss}
			\end{equation}
			
			\subsection{Post-processing Accumulator}
			In our framework, the accumulator functions as a post-processing step designed to enhance the temporal consistency of the predicted action phases. Raw model predictions often exhibit short-term fluctuations, which can lead to spurious or unstable phase labels. To address this, we introduce an accumulator-based smoothing mechanism. Since the action phases under consideration are temporally extended and follow a strictly sequential order without repetitions or regressions, we constrain transitions to occur only when the predicted phase directly follows the current one in the predefined sequence. Each time such a legal transition is detected, an accumulator is initialized and incremented with consecutive supporting predictions. Once the accumulator surpasses a predefined threshold, the transition is confirmed, and the system advances to the next phase. This strategy reduces the impact of noisy frame-level predictions, enforces sequential validity, and ensures that only sustained evidence is sufficient to trigger a phase change.
			
			\subsection{Collaborative Platform and Continual Learning}
\begin{figure*}[!t]
				\centering
				\includegraphics[width=\columnwidth]{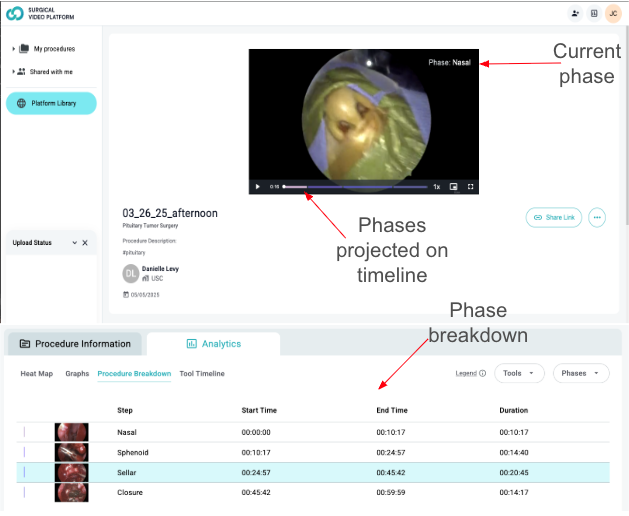}	
				\caption{An overview of the surgical online platform.}
				\label{fig:svp}
			\end{figure*}

			To enable real-world scalability, all system components are integrated into an online platform SVP through which surgeons can upload operative videos, receive automated phase analysis, and contribute to a growing dataset. As new data are collected, the platform supports continual model refinement, enabling incremental learning and improved generalization. The user interface of the online platform is displayed in Figure \ref{fig:svp}.
			
			\section{Experiments and Results}
			\label{sec:exp}
			\subsection{Data Curation}
			A total of 81 pituitary tumor surgery videos were collected from four institutions and the PitVis dataset \citep{das2025pitvis}. The average procedure duration was approximately 98 minutes, comprising 34 complete surgeries and 47 incomplete ones. All videos were anonymized prior to submission to the platform. The dataset encompassed diverse cases, including variations in zoom levels, surgical phases, and rare or atypical scenarios. Each video was recorded via an endoscopic system at 30 or 60 frames per second (fps). The video resolution varies across institutions due to different endoscopic systems used. SVP converts all videos to 1920 $\times $720 at 15fps. This study received ethical approval from the respective institutional review boards of the participating hospitals.
			
			For standardized model training, the videos were downsampled to 1 fps and resized to a resolution of $224\times 224$ pixels. The dataset was divided into training (62 videos; 174,267 frames), validation (8 videos; 28,989 frames), and testing (11 videos; 46,528 frames) subsets. Notably, the test set included three procedures performed by a surgeon not present in either the training or validation sets, thereby enabling evaluation of model generalization to unseen operators. During data loading, a custom augmentation step was applied to remove the black borders of the endoscopic view before resizing, thereby reducing information loss from resizing padded frames.
			
\begin{figure}[!t]
				\centering
				\includegraphics[width=\columnwidth]{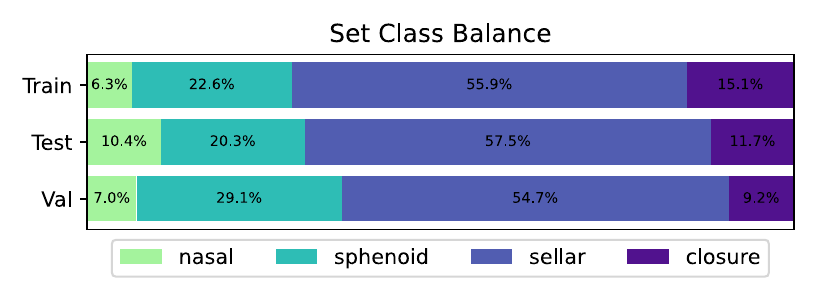}	
				\caption{Phase class distribution in the pituitary tumor surgery dataset. }
				\label{fig:pts_class_distrib}
			\end{figure}

\begin{figure}[!t]
				\centering
				\includegraphics[width=0.8\columnwidth]{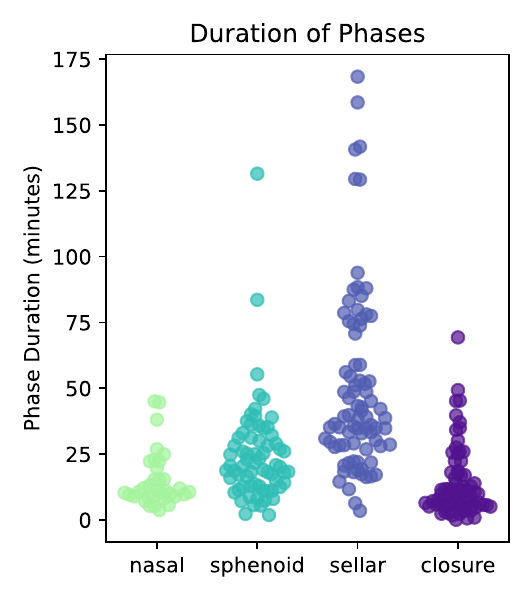}	
				\caption{The duration statistic of each phase class. }
				\label{fig:duration_stats}
			\end{figure}

\begin{figure}[!t]
				\centering
				\includegraphics[width=\columnwidth]{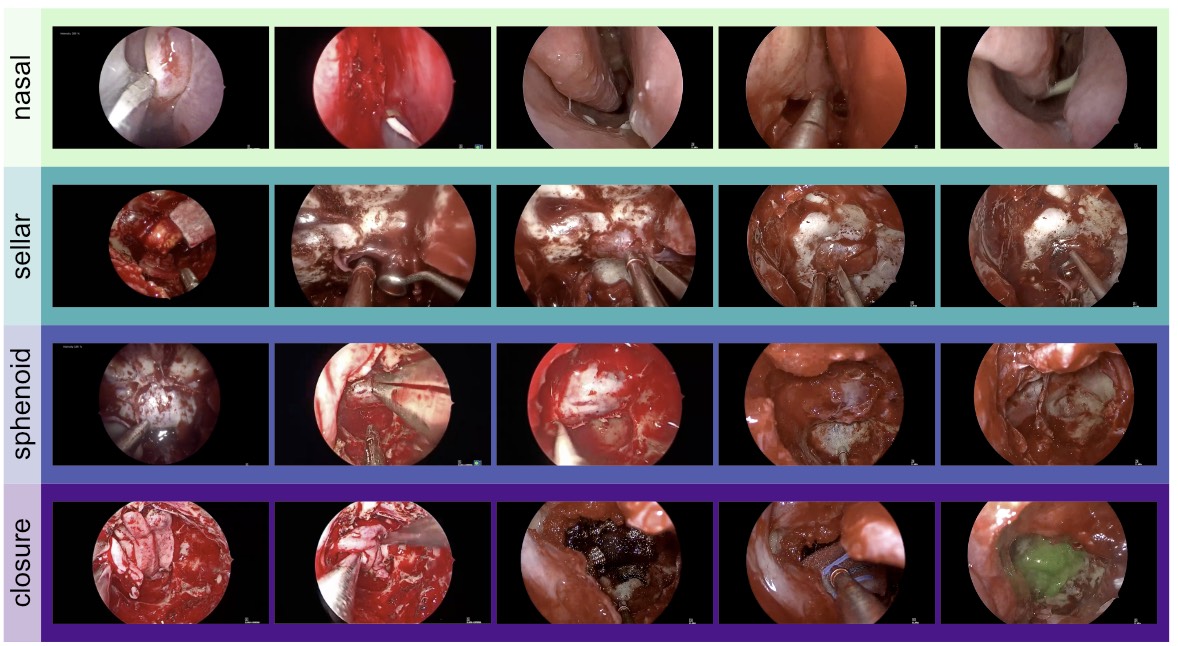}
				\caption{The characteristic frames of each phase class. }
				\label{fig:characteristic_frames}
			\end{figure}

			To enhance model generalizability across different surgical techniques and styles, the surgeries were categorized into four broad procedural phase classes based on the pituitary society expert Delphi consensus \citep{marcus2021pituitary}: nasal, representing navigation through the nasal corridor; sphenoid, indicating the drilling of an opening to access the tumor; sellar, denoting tumor resection; and closure, corresponding to reconstruction and completion of the procedure. The annotations were made on the platform by two surgeons. Ex-situ frames are removed, as well as very blurry frames, detected using a laplacian filter. The frames are further cropped to the size of the endoscope and resized. The phase class distributions are illustrated in Figure \ref{fig:pts_class_distrib}, the statistic of each phase class duration is shown in Figure \ref{fig:duration_stats}, and characteristic frames in each phase are demonstrated in Figure \ref{fig:characteristic_frames}.

			\subsection{Model Training}
			
			For the ResNet pretraining stage, the model was trained for 100 epochs using a learning rate of 0.075 and a batch size of 256. During fine-tuning, the learning rate was reduced to $1 \times 10^{-5}$, with a batch size of 128, and the model was trained for 15 epochs with eight H100SXM GPUs on Chicago Booth's Pythia Supercomputing Cluster. Subsequently, the MSTCN++ was also trained for 100 epochs with a learning rate of $1 \times 10^{-5}$ and a batch size of 1. The model was implemented in PyTorch \citep{ansel2024pytorch} and trained on an AWS EC2 instance equipped with one NVIDIA T4 GPU. Training employed the AdamW optimizer with the learning rate decaying according to a cosine annealing schedule. The training images were shuffled to ensure that all procedural phases were represented in each batch. Early stopping was applied, terminating training if the validation loss increased for three consecutive epochs. The fine-tuning of the ResNet model required five hours of training, whereas the subsequent training of the MS-TCN model was completed in one hour.

			\subsection{Evaluation Metrics}
			The performance of the surgical phase recognition model was assessed using four standard classification metrics: accuracy, precision, recall, and F1-score.
			\begin{itemize}
				\item Accuracy quantifies the proportion of correctly classified frames among all frames in the dataset and serves as an overall indicator of model performance. However, as accuracy can be influenced by class imbalance, additional metrics such as precision and recall are needed to provide a more comprehensive evaluation.
				\item Precision measures the proportion of correctly predicted instances of a given phase relative to all instances predicted as that phase, reflecting the model’s ability to avoid false positives.
				\item Recall represents the proportion of correctly predicted instances of a phase relative to all ground-truth instances of that phase, indicating the model’s capability to minimize false negatives.
				\item F1-score is the harmonic mean of precision and recall, providing a balanced metric that accounts for both types of classification errors.
			\end{itemize}
			All metrics were computed on a per-class basis and averaged to obtain macro-level scores, ensuring equal weight for each surgical phase regardless of class frequency.
			
			\subsection{Results}
			\subsubsection{Comparison with Existing Method}
\begin{figure}[!t]
				\centering
				\includegraphics[width=\columnwidth]{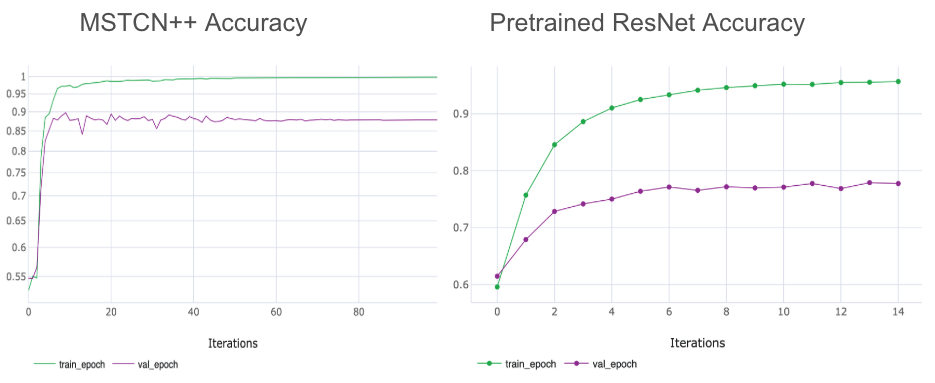}	
				\caption{Network performance over epochs.}
				\label{fig:epochs}
			\end{figure}

			The proposed model was evaluated on a held-out test set comprising 11 complete procedural videos. The evolution of the training loss across epochs is depicted in Figure \ref{fig:epochs}, illustrating the model’s stable convergence. Quantitative results are summarized in Table \ref{tab:phase_metrics}, alongside comparisons to the only other publicly reported pituitary surgery workflow analysis approaches \citep{das2025pitvis}. The latter set of methods report performance primarily in terms of F1-score and Edit score; however, because Edit score is less interpretable and rarely adopted as a primary metric in surgical phase recognition literature, we restrict our comparison to widely accepted classification metrics: precision, recall, F1-score, and overall accuracy.
			
			From Table \ref*{tab:phase_metrics}, it is evident that our method substantially outperforms existing baselines, achieving an overall F1-score of 87.91, which represents an absolute improvement of 44\% over the strongest prior approach (CITI, F1 = 61.10), with the highest F1 observed in the Sellar phase (92.54) and strong results in both Nasal and Closure phases despite their relatively short durations. These results indicate that the proposed self-supervised pre-training, focal-loss based fine-tuning, and MS-TCN++ temporal modeling pipeline are effective in capturing both spatial and temporal characteristics of endoscopic pituitary tumor surgeries, even in phases with class imbalance and frequent transitions.
			
\begin{figure}[!t]
				\centering
				\includegraphics[width=\columnwidth]{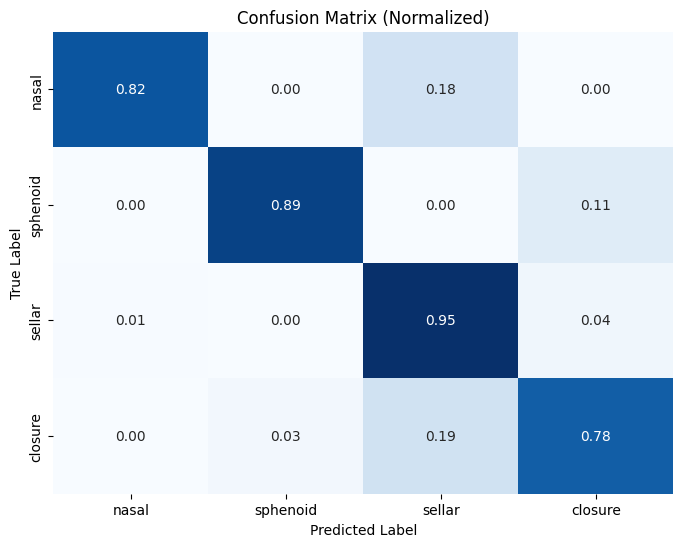}	
				\caption{Normalized confusion matrix for the phase classification.}
				\label{fig:confusion}
			\end{figure}

\begin{figure}[!t]
				\centering
				\includegraphics[width=\columnwidth]{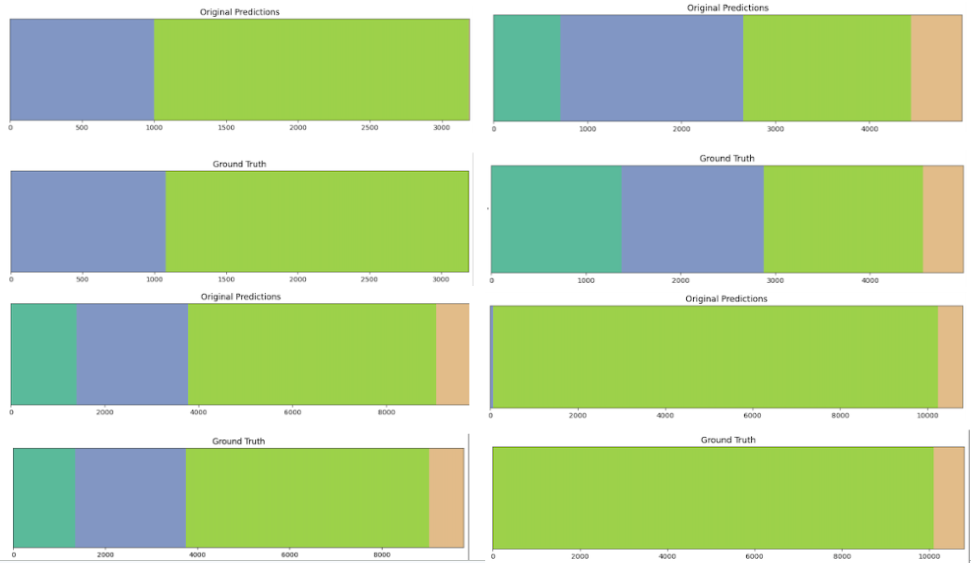}	
				\caption{An example comparison between the phase recognition result and ground truth.}
				\label{fig:ribbon}
			\end{figure}

			The confusion matrix in Figure \ref{fig:confusion} highlights that the majority of misclassifications occur between the sellar and closure phases, closure and sellar phase, which share similar visual characteristics in the early surgical context. An example qualitative comparison between predicted phase sequences and ground truth annotations is shown in Figure \ref{fig:ribbon}, further illustrating the model’s ability to produce temporally coherent predictions with minimal fragmentation.
			
			\begin{table}[t!]
				\centering
				\caption{SurgPhase online platform performance on pituitary tumor surgery phase recognition. }
				\label{tab:phase_metrics}
				\begin{tabular}{|c|c|c|c|c|c|}
					\Xhline{0.2ex}
					Method    &Phase        & Pre 	&Rec    & F1 & Acc  \\
					\hline
					\multirow{4}{*}{\makecell{SurgPhase\\(Our method)}}   & Nasal		& 94.22&88.75&91.40	& \multirow{4}{*}{90.00} \\
					\cline{2-5} 
					&Sphenoid  	&81.97&  78.27	&80.08	&	\\
					\cline{2-5} 
					&Sellar  	&89.99 &  95.24& 92.54 &\\
					\cline{2-5} 
					&Closure  	& 95.83 &80.81	&87.68&\\
					\hline
					CITI & \multirow{5}{*} {All phases} &-	&-&  61.10	& -	\\
					\cline{1-1} \cline{3-6} 
					UNI-ANDES & &-&- & 51.00	& -	\\
					\cline{1-1} \cline{3-6} 
					SK & &-&- & 41.20	& -	\\
					\cline{1-1} \cline{3-6} 
					SANO &  &-&- & 39.60	& -	\\
					\cline{1-1} \cline{3-6} 
					GMAI &  &-&- & 7.20	& -	\\
					\Xhline{0.2ex}
				\end{tabular}
			\end{table}
			
			\subsubsection{Ablation Experiments}
\begin{figure}[!t]
				\centering
				\includegraphics[width=0.8\columnwidth]{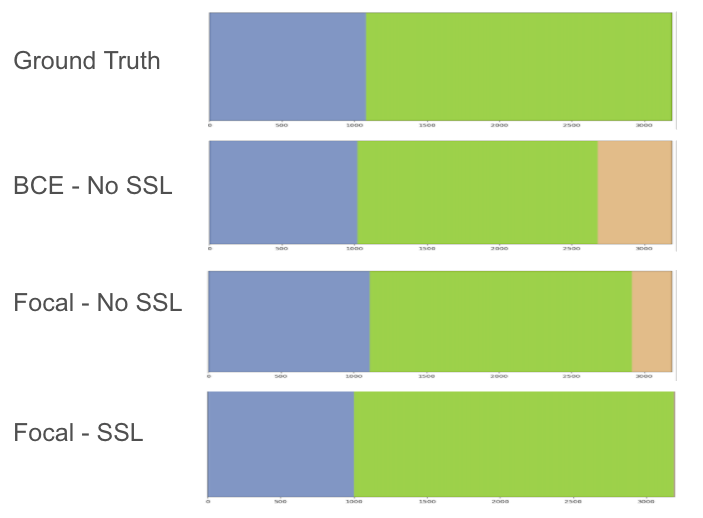}	
				\caption{An example comparison of pituitary tumor surgery action recognition results with different neural network configurations.}
				\label{fig:ablation_results}
			\end{figure}

			We evaluated our proposed model under three configurations: (1) baseline training with cross-entropy (BCE) loss without SSL; (2) focal loss without SSL; and (3) focal loss with SSL. The performance across surgical phases is reported in Table \ref*{tab:ablation} in terms of per-phase precision, recall, F1-score, and overall accuracy. Figure \ref{fig:ablation_results} is an example pituitary tumor surgery action recognition results comparison with these configuration.
			
			From the table, we observe that replacing BCE loss with focal loss yields a substantial improvement in overall accuracy, from 78.00\% to 81.00\%. This gain is particularly pronounced in phases with severe less class representation, such as the Nasal and Closure phases, where focal loss mitigates the dominance of majority-class samples by down-weighting well-classified instances and focusing gradient updates on harder, minority-class examples. In contrast, BCE loss tends to bias the decision boundaries toward overrepresented phases, leading to low recall in underrepresented ones. This is further supported by the observation in Figure \ref{fig:ablation_results}, which demonstrates that recognition of the underrepresented closure class fails when BCE is employed. 
			
			Incorporating SSL pretraining on 251 unlabeled procedures further improves accuracy to 90.00\%, with marked gains in both recall and precision across all phases.
			
			These results confirm that our SSL + focal loss combination substantially improves phase recognition performance, especially for phases with limited labeled data, and is well-suited for deployment in our online surgical analytics platform where robustness across all phases is critical.

			\begin{table}[t!]
				\centering
				\caption{Ablation study. Comparison of performance without focal loss and self-supervised learning.}
				\label{tab:ablation}
				\resizebox{0.98\linewidth}{!}{
					\begin{tabular}{|c|c|c|c|c|c|c|c|c|}
						\Xhline{0.2ex}
						Method    &Phase        & Pre 	&Rec    & F1 & Acc  & Pre & Rec & F1\\
						\hline

						\multirow{4}{*}{\makecell{SurgPhase\\ w/ BCE\\ w/o SSL}}   
						& Nasal		& 81.73 & 70.83 & 75.89
						& \multirow{4}{*}{78.12} 
						& \multirow{4}{*}{76.05}
						& \multirow{4}{*}{80.10}
						& \multirow{4}{*}{77.34}
						\\ \cline{2-5}
						
						& Sphenoid  	& 73.13 & 94.78 & 82.56
						& \multicolumn{1}{c|}{} & \multicolumn{1}{c|}{} & \multicolumn{1}{c|}{} & \multicolumn{1}{c|}{}
						\\ \cline{2-5}
						
						& Sellar  		& 91.65 & 76.25 & 83.25
						& \multicolumn{1}{c|}{} & \multicolumn{1}{c|}{} & \multicolumn{1}{c|}{} & \multicolumn{1}{c|}{}
						\\ \cline{2-5}
						
						& Closure  	& 55.65 & 77.89 & 64.92
						& \multicolumn{1}{c|}{} & \multicolumn{1}{c|}{} & \multicolumn{1}{c|}{} & \multicolumn{1}{c|}{}
						\\
						\hline
						
						\multirow{4}{*}{\makecell{SurgPhase\\ w/ focal loss \\ w/o SSL}}   
						& Nasal		&  92.88 & 42.62 & 58.43 
						& \multirow{4}{*}{81.22} 
						& \multirow{4}{*}{83.37}
						& \multirow{4}{*}{74.84} 
						& \multirow{4}{*}{76.01} 
						\\ \cline{2-5}
						
						& Sphenoid  	& 90.34 & 90.30 & 90.32 
						& \multicolumn{1}{c|}{} & \multicolumn{1}{c|}{} & \multicolumn{1}{c|}{} & \multicolumn{1}{c|}{}
						\\ \cline{2-5}
						
						& Sellar  		& 84.14 & 88.59 & 86.31
						& \multicolumn{1}{c|}{} & \multicolumn{1}{c|}{} & \multicolumn{1}{c|}{} & \multicolumn{1}{c|}{}
						\\ \cline{2-5}
						
						& Closure  	& 65.21 & 75.74 & 70.08
						& \multicolumn{1}{c|}{} & \multicolumn{1}{c|}{} & \multicolumn{1}{c|}{} & \multicolumn{1}{c|}{}
						\\
						\hline
						
						\multirow{4}{*}{\makecell{SurgPhase\\ w/ focal loss\\ w/ SSL}}   
						& Nasal		& 95.82 & 81.53 & 88.10
						& \multirow{4}{*}{90.00} 
						& \multirow{4}{*}{91.25}
						& \multirow{4}{*}{86.02}
						& \multirow{4}{*}{88.30}
						\\ \cline{2-5}
						
						& Sphenoid  	& 94.23 & 88.75 & 91.41
						& \multicolumn{1}{c|}{} & \multicolumn{1}{c|}{} & \multicolumn{1}{c|}{} & \multicolumn{1}{c|}{}
						\\ \cline{2-5}
						
						& Sellar  		& 90.11 & 95.25 & 92.61
						& \multicolumn{1}{c|}{} & \multicolumn{1}{c|}{} & \multicolumn{1}{c|}{} & \multicolumn{1}{c|}{}
						\\ \cline{2-5}
						
						& Closure  	& 82.02 & 78.27 & 80.10
						& \multicolumn{1}{c|}{} & \multicolumn{1}{c|}{} & \multicolumn{1}{c|}{} & \multicolumn{1}{c|}{}
						\\
						\Xhline{0.2ex}
					\end{tabular}
				}
			\end{table}
			
			\section{Discussion}
			\label{sec:dis}
			Surgical phase recognition has become a pivotal component of intelligent surgical systems, offering the potential to enhance intraoperative decision support, automate postoperative video review, and enable large-scale surgical skill assessment. While this research has gained considerable traction in domains such as laparoscopic cholecystectomy \citep{twinanda2016single,madani2022artificial}, cataract surgery \citep{fang2022global}, and colorectal resection \citep{endonet}, relatively little work has been devoted to pituitary tumor surgery, which is an intricate procedure characterized by small working spaces, complex anatomy, and high variability in surgical workflow. This scarcity of annotated pituitary datasets and research works underscores both the novelty and the great value of our contribution. By providing a robust method for this PTS phase recognition, coupled with a collaborative online platform for continuous data collection and surgeon engagement, our work addresses an unmet need in the neurosurgical community.
			
			Our experimental results demonstrate that combining SSL with focal loss yields substantial performance improvements over conventional supervised BCE training. The baseline BCE model achieved an overall accuracy of 78\%, primarily due to severe class imbalance and phase boundary ambiguity. Replacing BCE with focal loss improved accuracy to 81\%, confirming its efficacy in focusing learning on minority-class and hard-to-classify samples. Incorporating SSL pretraining further boosted performance to 90.00\%, with particularly large gains in the Nasal and Sellar phases. These results align with findings from other domains, where SSL-derived features improve downstream phase classification, especially in data-limited settings.
			
			Compared with other state-of-the-art surgical phase recognition approaches, including TCN-based methods \citep{farha2019ms} and transformer-based pipelines \citep{yang2024surgformer, perez2024must}, our framework achieves competitive or superior accuracy despite operating in a more complex surgical setting with fewer labeled examples. Transformer-based approaches often excel in modeling long-range dependencies but typically require larger labeled datasets and incur high inference latency, making them less suitable for our online analysis requirement where surgeons upload videos and expect near real-time feedback. Our MS-TCN++ backbone offers a favorable trade-off between temporal modeling capacity and computational efficiency, enabling integration into a web-based collaborative platform without prohibitive resource demands.
			
			Importantly, our findings suggest that SSL not only boosts performance in traditional train–test splits but also improves generalization to procedures with different numbers of phases or varying start and end points, a key capability for real-world deployment where surgical workflows are rarely uniform.
			
			Beyond technical performance, the interactive online platform magnifies the societal value of this work. By enabling surgeons worldwide to upload procedure videos, receive automated phase analysis, and access objective metrics, the platform supports surgical education, self-assessment, and peer-to-peer learning. Trainee surgeons can review annotated timelines to better understand workflow sequencing and decision-making patterns, while experienced surgeons can evaluate and refine their techniques based on quantitative feedback. Moreover, in low- and middle-income countries, where access to specialized neurosurgeons is limited, such a system can serve as a remote mentoring and quality assurance tool, bridging geographical gaps and democratizing surgical expertise. Over time, the continuous accumulation of annotated data will not only improve the model’s accuracy but also create an unprecedented global knowledge base of pituitary tumor surgeries, fostering large-scale surgical outcomes research.
			
			\subsection{Limitations}
			
			First, while the model generalizes better to varied procedures, we observed a modest decrease in accuracy for some individual videos that deviates from the majority video patterns. Second, phase boundaries, particularly between Sellar and Sphenoid, remain challenging to detect due to visual similarity, which could be mitigated by incorporating surgical tool usage information with object detection model. Third, sparsely populated Closure phases are harder to identify reliably, suggesting that better sampling or targeted augmentation strategies are needed. Fourth, some frames within a phase may be uninformative or misleading, suggesting improved frame selection could enhance phase discriminability. Finally, while the SSL-pretrained ResNet encoder performs strongly, the MS-TCN++ temporal aggregation stage appears to have more room for improvement, indicating that future work should explore more adaptive or hybrid temporal models.
			
			Overall, this study demonstrates that combining SSL, focal loss, and robust temporal modeling offers a practical and high-performing solution for pituitary tumor surgery phase recognition. Beyond its technical contributions, the integration into an online surgeon collaboration platform positions the system for real-world clinical adoption and iterative refinement, paving the way for broader applications in neurosurgical workflow analysis.
			
			\section{Conclusion}
			\label{sec:con}
			We proposed a surgical phase recognition framework tailored to pituitary tumor surgery, integrating SSL-pretrained ResNet-50 encoding, focal loss for class imbalance, and a modified MS-TCN++ for temporal modeling. Trained on a combination of unlabeled and labeled surgical videos, our method achieved 90.00\% accuracy, outperforming baselines and showing strong robustness to workflow variability. By embedding this model into a collaborative online platform, we enable automated, objective, and scalable surgical workflow analysis, addressing both technical and clinical needs. Future work will focus on improving phase boundary detection, optimizing temporal aggregation, and expanding multimodal integration to further enhance accuracy and clinical utility.
			
			\section*{CRediT authorship contribution statement}
			\textbf{Yan Meng:} Writing -review \& editing, Writing - original draft, Methodology. \textbf{Jack Cook:} Implementation, Writing-review. \textbf{Kaan Duman:} Implementation support. \textbf{Margaux Masson-Forsythe:} Supervision, Writing-review, Implementation support. \textbf{X.Y. Han:} Resources, Implementation support - SSL, Writing - review. 
			\textbf{Shauna Otto and Dhiraj Pangal} Writing - review, illustration.
			\textbf{Jonathan Chainey and Dr Ruth Lau} Writing - review.
			\textbf{Daniel A. Donoho:} Writing - review, Resources, Supervision, Funding acquisition. 
			\textbf{Danielle Levy, Gabriel Zada, Sébastien Froelich, Juan Fernandez-Miranda, Mike Chang:} Data acquisition. 
			\section*{ Declaration of competing interest}
			The authors declare that they have no known competing financial interests or personal relationships that could have appeared to influence the work reported in this paper.
			
			\section*{Funding}
			The project is funded by Surgical Data Science Collective (SDSC). X.Y. Han is funded by research support from the Booth School of Business at the University of Chicago with high-performance computing resources from Chicago Booth's Pythia Supercomputer Cluster.
			
			\section*{Acknowledgments}
			SDSC engineering team: Ahmed Amin, Ameya Mangalvedhekar, Sumeyra O. Kaplan, Andrew Rama, Dor Spitzer, Tristam MacDonald, Yajur Sehra.
			
			
			\bibliographystyle{elsarticle-harv} 
			\bibliography{ref}

@inproceedings{he2016deep,
	title={Deep residual learning for image recognition},
	author={He, Kaiming and Zhang, Xiangyu and Ren, Shaoqing and Sun, Jian},
	booktitle={Proceedings of the IEEE conference on computer vision and pattern recognition},
	pages={770--778},
	year={2016}
}

@inproceedings{chen2020simple,
	title={A simple framework for contrastive learning of visual representations},
	author={Chen, Ting and Kornblith, Simon and Norouzi, Mohammad and Hinton, Geoffrey},
	booktitle={International conference on machine learning},
	pages={1597--1607},
	year={2020},
	organization={PmLR}
}

@inproceedings{lin2017focal,
	title={Focal loss for dense object detection},
	author={Lin, Tsung-Yi and Goyal, Priya and Girshick, Ross and He, Kaiming and Doll{\'a}r, Piotr},
	booktitle={Proceedings of the IEEE international conference on computer vision},
	pages={2980--2988},
	year={2017}
}

@inproceedings{farha2019ms,
	title={Ms-tcn: Multi-stage temporal convolutional network for action segmentation},
	author={Farha, Yazan Abu and Gall, Jurgen},
	booktitle={Proceedings of the IEEE/CVF conference on computer vision and pattern recognition},
	pages={3575--3584},
	year={2019}
}

@article{das2025pitvis,
	title={Pitvis-2023 challenge: Workflow recognition in videos of endoscopic pituitary surgery},
	author={Das, Adrito and Khan, Danyal Z and Psychogyios, Dimitrios and Zhang, Yitong and Hanrahan, John G and Vasconcelos, Francisco and Pang, You and Chen, Zhen and Wu, Jinlin and Zou, Xiaoyang and others},
	journal={Medical Image Analysis},
	pages={103716},
	year={2025},
	publisher={Elsevier}
}

@inproceedings{czempiel2020tecno,
	title={Tecno: Surgical phase recognition with multi-stage temporal convolutional networks},
	author={Czempiel, Tobias and Paschali, Magdalini and Keicher, Matthias and Simson, Walter and Feussner, Hubertus and Kim, Seong Tae and Navab, Nassir},
	booktitle={Medical Image Computing and Computer Assisted Intervention--MICCAI 2020: 23rd International Conference, Lima, Peru, October 4--8, 2020, Proceedings, Part III 23},
	pages={343--352},
	year={2020},
	organization={Springer}
}

@article{ramesh2023dissecting,
	title={Dissecting self-supervised learning methods for surgical computer vision},
	author={Ramesh, Sanat and Srivastav, Vinkle and Alapatt, Deepak and Yu, Tong and Murali, Aditya and Sestini, Luca and Nwoye, Chinedu Innocent and Hamoud, Idris and Sharma, Saurav and Fleurentin, Antoine and others},
	journal={Medical Image Analysis},
	volume={88},
	pages={102844},
	year={2023},
	publisher={Elsevier}
}

@inproceedings{yang2024surgformer,
	title={Surgformer: Surgical transformer with hierarchical temporal attention for surgical phase recognition},
	author={Yang, Shu and Luo, Luyang and Wang, Qiong and Chen, Hao},
	booktitle={International Conference on Medical Image Computing and Computer-Assisted Intervention},
	pages={606--616},
	year={2024},
	organization={Springer}
}

@article{wong2024collaborative,
	title={Collaborative Human--Computer Vision Operative Video Analysis Algorithm for Analyzing Surgical Fluency and Surgical Interruptions in Endonasal Endoscopic Pituitary Surgery: Cohort Study},
	author={Wong, Chia-En and Chen, Pei-Wen and Hsu, Heng-Jui and Cheng, Shao-Yang and Fan, Chen-Che and Chen, Yen-Chang and Chiu, Yi-Pei and Lee, Jung-Shun and Liang, Sheng-Fu},
	journal={Journal of medical Internet research},
	volume={26},
	pages={e56127},
	year={2024},
	publisher={JMIR Publications Toronto, Canada}
}

@article{twinanda2016single,
	title={Single-and multi-task architectures for surgical workflow challenge at M2CAI 2016},
	author={Twinanda, Andru P and Mutter, Didier and Marescaux, Jacques and de Mathelin, Michel and Padoy, Nicolas},
	journal={arXiv preprint arXiv:1610.08844},
	year={2016}
}

@article{ramesh2021multi,
	title={Multi-task temporal convolutional networks for joint recognition of surgical phases and steps in gastric bypass procedures},
	author={Ramesh, Sanat and Dall’Alba, Diego and Gonzalez, Cristians and Yu, Tong and Mascagni, Pietro and Mutter, Didier and Marescaux, Jacques and Fiorini, Paolo and Padoy, Nicolas},
	journal={International journal of computer assisted radiology and surgery},
	volume={16},
	number={7},
	pages={1111--1119},
	year={2021},
	publisher={Springer}
}

@article{marcus2021pituitary,
	title={Pituitary society expert Delphi consensus: operative workflow in endoscopic transsphenoidal pituitary adenoma resection},
	author={Marcus, Hani J and Khan, Danyal Z and Borg, Anouk and Buchfelder, Michael and Cetas, Justin S and Collins, Justin W and Dorward, Neil L and Fleseriu, Maria and Gurnell, Mark and Javadpour, Mohsen and others},
	journal={Pituitary},
	volume={24},
	number={6},
	pages={839--853},
	year={2021},
	publisher={Springer}
}

@inproceedings{ansel2024pytorch,
	title={Pytorch 2: Faster machine learning through dynamic python bytecode transformation and graph compilation},
	author={Ansel, Jason and Yang, Edward and He, Horace and Gimelshein, Natalia and Jain, Animesh and Voznesensky, Michael and Bao, Bin and Bell, Peter and Berard, David and Burovski, Evgeni and others},
	booktitle={Proceedings of the 29th ACM International Conference on Architectural Support for Programming Languages and Operating Systems, Volume 2},
	pages={929--947},
	year={2024}
}

@inproceedings{carreira2017quo,
	title={Quo vadis, action recognition? a new model and the kinetics dataset},
	author={Carreira, Joao and Zisserman, Andrew},
	booktitle={proceedings of the IEEE Conference on Computer Vision and Pattern Recognition},
	pages={6299--6308},
	year={2017}
}

@inproceedings{tran2018closer,
	title={A closer look at spatiotemporal convolutions for action recognition},
	author={Tran, Du and Wang, Heng and Torresani, Lorenzo and Ray, Jamie and LeCun, Yann and Paluri, Manohar},
	booktitle={Proceedings of the IEEE conference on Computer Vision and Pattern Recognition},
	pages={6450--6459},
	year={2018}
}

@inproceedings{feichtenhofer2019slowfast,
	title={Slowfast networks for video recognition},
	author={Feichtenhofer, Christoph and Fan, Haoqi and Malik, Jitendra and He, Kaiming},
	booktitle={Proceedings of the IEEE/CVF international conference on computer vision},
	pages={6202--6211},
	year={2019}
}

@article{park2022multi,
	title={Multi-stage temporal convolutional network with moment loss and positional encoding for surgical phase recognition},
	author={Park, Minyoung and Oh, Seungtaek and Jeong, Taikyeong and Yu, Sungwook},
	journal={Diagnostics},
	volume={13},
	number={1},
	pages={107},
	year={2022},
	publisher={MDPI}
}

@article{fang2022global,
	title={Global--local multi-stage temporal convolutional network for cataract surgery phase recognition},
	author={Fang, Lixin and Mou, Lei and Gu, Yuanyuan and Hu, Yan and Chen, Bang and Chen, Xu and Wang, Yang and Liu, Jiang and Zhao, Yitian},
	journal={BioMedical Engineering OnLine},
	volume={21},
	number={1},
	pages={82},
	year={2022},
	publisher={Springer}
}

@inproceedings{lea2017temporal,
	title={Temporal convolutional networks for action segmentation and detection},
	author={Lea, Colin and Flynn, Michael D and Vidal, Rene and Reiter, Austin and Hager, Gregory D},
	booktitle={proceedings of the IEEE Conference on Computer Vision and Pattern Recognition},
	pages={156--165},
	year={2017}
}

@article{li2020ms,
	author={Shi-Jie Li and Yazan AbuFarha and Yun Liu and Ming-Ming Cheng and Juergen Gall},
	journal={IEEE Transactions on Pattern Analysis and Machine Intelligence}, 
	title={MS-TCN++: Multi-Stage Temporal Convolutional Network for Action Segmentation}, 
	year={2020},
	volume={},
	number={},
	pages={1-1},
	doi={10.1109/TPAMI.2020.3021756},
}

@article{yi2021asformer,
	title={Asformer: Transformer for action segmentation},
	author={Yi, Fangqiu and Wen, Hongyu and Jiang, Tingting},
	journal={arXiv preprint arXiv:2110.08568},
	year={2021}
}

@inproceedings{czempiel2021opera,
	title={Opera: Attention-regularized transformers for surgical phase recognition},
	author={Czempiel, Tobias and Paschali, Magdalini and Ostler, Daniel and Kim, Seong Tae and Busam, Benjamin and Navab, Nassir},
	booktitle={International conference on medical image computing and computer-assisted intervention},
	pages={604--614},
	year={2021},
	organization={Springer}
}

@inproceedings{gao2021trans,
	title={Trans-svnet: Accurate phase recognition from surgical videos via hybrid embedding aggregation transformer},
	author={Gao, Xiaojie and Jin, Yueming and Long, Yonghao and Dou, Qi and Heng, Pheng-Ann},
	booktitle={International conference on medical image computing and computer-assisted intervention},
	pages={593--603},
	year={2021},
	organization={Springer}
}

@article{liu2025lovit,
	title={Lovit: Long video transformer for surgical phase recognition},
	author={Liu, Yang and Boels, Maxence and Garcia-Peraza-Herrera, Luis C and Vercauteren, Tom and Dasgupta, Prokar and Granados, Alejandro and Ourselin, Sebastien},
	journal={Medical Image Analysis},
	volume={99},
	pages={103366},
	year={2025},
	publisher={Elsevier}
}

@inproceedings{perez2024must,
	title={Must: Multi-scale t ransformers for surgical phase recognition},
	author={P{\'e}rez, Alejandra and Rodr{\'\i}guez, Santiago and Ayobi, Nicol{\'a}s and Aparicio, Nicol{\'a}s and Dessevres, Eug{\'e}nie and Arbel{\'a}ez, Pablo},
	booktitle={International Conference on Medical Image Computing and Computer-Assisted Intervention},
	pages={422--432},
	year={2024},
	organization={Springer}
}

@article{batic2024endovit,
	title={EndoViT: pretraining vision transformers on a large collection of endoscopic images},
	author={Bati{\'c}, Dominik and Holm, Felix and {\"O}zsoy, Ege and Czempiel, Tobias and Navab, Nassir},
	journal={International Journal of Computer Assisted Radiology and Surgery},
	volume={19},
	number={6},
	pages={1085--1091},
	year={2024},
	publisher={Springer}
}

@article{das2023automatic,
	title={Automatic generation of operation notes in endoscopic pituitary surgery videos using workflow recognition},
	author={Das, Adrito and Khan, Danyal Z and Hanrahan, John G and Marcus, Hani J and Stoyanov, Danail},
	journal={Intelligence-Based Medicine},
	volume={8},
	pages={100107},
	year={2023},
	publisher={Elsevier}
}

@article{nyangoh2023systematic,
	title={A systematic review of annotation for surgical process model analysis in minimally invasive surgery based on video},
	author={Nyangoh Timoh, Krystel and Huaulme, Arnaud and Cleary, Kevin and Zaheer, Myra A and Lavoue, Vincent and Donoho, Dan and Jannin, Pierre},
	journal={Surgical endoscopy},
	volume={37},
	number={6},
	pages={4298--4314},
	year={2023},
	publisher={Springer}
}

@article{nwoye2023cholectriplet2021,
	title={CholecTriplet2021: A benchmark challenge for surgical action triplet recognition},
	author={Nwoye, Chinedu Innocent and Alapatt, Deepak and Yu, Tong and Vardazaryan, Armine and Xia, Fangfang and Zhao, Zixuan and Xia, Tong and Jia, Fucang and Yang, Yuxuan and Wang, Hao and others},
	journal={Medical Image Analysis},
	volume={86},
	pages={102803},
	year={2023},
	publisher={Elsevier}
}

@article{al2019cataracts,
	title={CATARACTS: Challenge on automatic tool annotation for cataRACT surgery},
	author={Al Hajj, Hassan and Lamard, Mathieu and Conze, Pierre-Henri and Roychowdhury, Soumali and Hu, Xiaowei and Mar{\v{s}}alkait{\.e}, Gabija and Zisimopoulos, Odysseas and Dedmari, Muneer Ahmad and Zhao, Fenqiang and Prellberg, Jonas and others},
	journal={Medical image analysis},
	volume={52},
	pages={24--41},
	year={2019},
	publisher={Elsevier}
}

@article{endonet,
	author = {Twinanda, Andru and Shehata, Sherif and Mutter, Didier and Marescaux, Jacques and De Mathelin, Michel and Padoy, Nicolas},
	year = {2016},
	month = {02},
	title = {EndoNet: A Deep Architecture for Recognition Tasks on Laparoscopic Videos},
	volume = {36},
	journal = {IEEE Transactions on Medical Imaging},
	doi = {10.1109/TMI.2016.2593957}
}

@article{meara2015global,
	title={Global Surgery 2030: evidence and solutions for achieving health, welfare, and economic development},
	author={Meara, John G and Leather, Andrew JM and Hagander, Lars and Alkire, Blake C and Alonso, Nivaldo and Ameh, Emmanuel A and Bickler, Stephen W and Conteh, Lesong and Dare, Anna J and Davies, Justine and others},
	journal={The lancet},
	volume={386},
	number={9993},
	pages={569--624},
	year={2015},
	publisher={Elsevier}
}

@article{alkire2015global,
	title={Global access to surgical care: a modelling study},
	author={Alkire, Blake C and Raykar, Nakul P and Shrime, Mark G and Weiser, Thomas G and Bickler, Stephen W and Rose, John A and Nutt, Cameron T and Greenberg, Sarah LM and Kotagal, Meera and Riesel, Johanna N and others},
	journal={The Lancet Global Health},
	volume={3},
	number={6},
	pages={e316--e323},
	year={2015},
	publisher={Elsevier}
}

@article{madani2022artificial,
	title={Artificial intelligence for intraoperative guidance: using semantic segmentation to identify surgical anatomy during laparoscopic cholecystectomy},
	author={Madani, Amin and Namazi, Babak and Altieri, Maria S and Hashimoto, Daniel A and Rivera, Angela Maria and Pucher, Philip H and Navarrete-Welton, Allison and Sankaranarayanan, Ganesh and Brunt, L Michael and Okrainec, Allan and others},
	journal={Annals of surgery},
	volume={276},
	number={2},
	pages={363--369},
	year={2022},
	publisher={LWW}
}

@article{hashimoto2018artificial,
	title={Artificial intelligence in surgery: promises and perils},
	author={Hashimoto, Daniel A and Rosman, Guy and Rus, Daniela and Meireles, Ozanan R},
	journal={Annals of surgery},
	volume={268},
	number={1},
	pages={70--76},
	year={2018},
	publisher={LWW}
}

@article{khalfaoui2021dilated,
	title={Dilated convolution with learnable spacings},
	author={Khalfaoui-Hassani, Ismail and Pellegrini, Thomas and Masquelier, Timoth{\'e}e},
	journal={arXiv preprint arXiv:2112.03740},
	year={2021}
}

@article{fagan2014open,
	title={Open access atlas of otolaryngology, head \& neck operative surgery},
	author={Fagan, Johan},
	journal={online},
	year={2014}
}

@article{sharma2016endoscopic,
	title={Endoscopic pituitary surgery: Techniques, tips and tricks, nuances, and complication avoidance},
	author={Sharma, Bhawani Shanker and Sawarkar, Dattaraj Paramanand and Suri, Ashish},
	journal={Neurology India},
	volume={64},
	number={4},
	pages={724--736},
	year={2016},
	publisher={Medknow}
}

@article{meng2025ai,
	title={AI-Driven Evaluation of Surgical Skill via Action Recognition},
	author={Meng, Yan and Donoho, Daniel A and Altshuler, Marcelle and Arnaout, Omar},
	journal={arXiv preprint arXiv:2512.24411},
	year={2025}
}

@article{vaswani2017attention,
	title={Attention is all you need},
	author={Vaswani, Ashish and Shazeer, Noam and Parmar, Niki and Uszkoreit, Jakob and Jones, Llion and Gomez, Aidan N and Kaiser, {\L}ukasz and Polosukhin, Illia},
	journal={Advances in neural information processing systems},
	volume={30},
	year={2017}
}
			
		\end{document}